\documentclass{article}

\usepackage{multirow}
\usepackage{multicol}
\usepackage{graphicx}
\usepackage{xspace}
\usepackage{amsmath}
\usepackage{adjustbox}
\usepackage{tcolorbox}
\usepackage{amssymb}
\usepackage{enumitem}
\usepackage{wrapfig}
\usepackage{xcolor}
\usepackage{subcaption}    %
\usepackage{float}
\usepackage{stfloats}
\usepackage{amsmath}
\usepackage{listings}
\usepackage{mdframed}
\usepackage{tcolorbox}
\usepackage{arydshln}
\usepackage{booktabs} 
\usepackage{color,soul}
\usepackage{makecell}
\usepackage[numbers]{natbib}
\usepackage{caption}
\usepackage{makecell}
\usepackage[accepted]{icml2025}
\usepackage[textsize=tiny]{todonotes}
\usepackage{hyperref}
\usepackage[capitalize,noabbrev]{cleveref}
\usepackage{arydshln}
\usepackage{hhline}
\usepackage{booktabs}

\tcbuselibrary{listings,breakable}

\definecolor{pastelblue}{RGB}{173,216,230}
\definecolor{pastelyellow}{RGB}{255,253,208}
\definecolor{pastelpink}{RGB}{255,209,220}
\definecolor{pastelgreen}{RGB}{176,226,172}
\definecolor{pastellavender}{RGB}{230,230,250}

\definecolor{NvidiaGreen}{RGB}{118, 185, 0}
\sethlcolor{red!15}

\icmltitlerunning{GenSelect: A Generative Approach to Best-of-$N$}

\begin{document}

\twocolumn[
\icmltitle{GenSelect: A Generative Approach to Best-of-$N$}



\icmlsetsymbol{equal}{*}

\begin{icmlauthorlist}
\icmlauthor{Shubham Toshniwal}{nv}
\icmlauthor{Ivan Sorokin}{nv}
\icmlauthor{Aleksander Ficek}{nv}
\icmlauthor{Ivan Moshkov}{nv}
\icmlauthor{Igor Gitman}{nv}
\end{icmlauthorlist}

\icmlaffiliation{nv}{NVIDIA}

\icmlcorrespondingauthor{Shubham Toshniwal}{stoshniwal@nvidia.com}

\icmlkeywords{Best-of-N, Test time scaling}

\vskip 0.3in
]
% \footnotetext{The second AI for MATH Workshop at the 42nd International Conference on Machine Learning, Vancouver, Canada.}

\newcommand{\deepseekr}{\texttt{DeepSeek-R1-0528}}



\printAffiliationsAndNotice{\icmlEqualContribution} % otherwise use the standard text.

\begin{abstract}

Generative reward models with parallel sampling have enabled effective test-time scaling for reasoning tasks. Current approaches employ pointwise scoring of individual solutions or pairwise comparisons. 
However, pointwise methods underutilize LLMs' comparative abilities, while pairwise methods scale inefficiently with larger sampling budgets.
We introduce GenSelect, where the LLM uses long reasoning to select the best solution among $N$ candidates. 
This leverages LLMs' comparative strengths while scaling efficiently across parallel sampling budgets. 
For math reasoning, we demonstrate that reasoning models, such as \texttt{QwQ} and \texttt{DeepSeek-R1-0528}, excel at GenSelect, outperforming existing scoring approaches with simple prompting. 
% We also demonstrate a simple RL recipe for training models on the GenSelect task.

% Reward models for tasks such as mathematical reasoning are often . We propose GenSelect, an extension of generative reward modeling for selecting the top-ranking candidate. 

\end{abstract}

\section{Introduction}

% This scaling for generation can be achieved by a combination of \emph{sequential} scaling with the use of reasoning models and \emph{parallel} sampling wherein multiple independent solutions are generated. 
% A key component of parallel sampling is selecting the top response among all candidates. 
% While typical strategies for scoring candidates involve the use of heuristics or discriminative reward models, recent work has shown that generative reward models (GenRMs) unlock yet another way for scaling test-time compute. 
% In fact, recent work has shown that verifiers/reward models can also benefit from extended reasoning. 

Test-time scaling has emerged as a powerful paradigm for enhancing the performance of large language models on reasoning tasks~\cite{openai2024openaio1card, deepseekai2025deepseekr1incentivizingreasoningcapability}. 
This scaling approach leverages two complementary mechanisms: \emph{sequential scaling}, which utilizes long reasoning models, and \emph{parallel sampling}, which produces multiple independent candidate solutions~\cite{snell2025scalingllmtesttimecompute}. 
Central to the parallel sampling strategy is the critical task of identifying the highest-quality response from the generated candidates. 
Typical candidate evaluation methods typically rely on rule-based methods~\cite{hassid2025dontoverthinkitpreferring} or discriminative reward models~\cite{cobbe2021trainingverifierssolvemath, liu2025acemathadvancingfrontiermath}.  
Recent work has demonstrated the efficacy of generative reward models (GenRMs), which offer another axis for leveraging test-time computation~\cite{zhang2025generativeverifiersrewardmodeling, mahan2024generativerewardmodels}. 
Moreover, recent findings suggest that these generative verification and reward models also benefit from extended reasoning~\cite{chen2025rmr1rewardmodelingreasoning, guo2025rewardreasoningmodel}.

% These generative reward models primarily operate in two modes: (a) \emph{Pointwise} in which the model scores each candidate output one at a time~\cite{zhang2025generativeverifiersrewardmodeling}, or (b) \emph{Pairwise} where the model compares two inputs at a time~\cite{zhao2025samplescrutinizescaleeffective}.  
% While pointwise GenRMs can be easily scaled with parallel sampling, they rely on the weak verification capabilities of LLMs, which can hinder their performance~\cite{tyen-etal-2024-llms, zhao2025samplescrutinizescaleeffective}.  
% On the other hand, pairwise GenRMs leverage the strengths of LLMs in comparison, but extending them 
% to larger parallel sampling budgets is non-trivial. For example, \citet{jiang-etal-2023-llm} perform $\mathcal{O}(N^2)$ pairwise comparisons to rank $N$ samples. 
% This cost is even more daunting for reasoning-based GenRMs. 

% While scores from pointwise approaches are easily ext

These generative reward models primarily operate through two distinct evaluation paradigms: \emph{pointwise evaluation}, where models assess individual candidate outputs in isolation~\cite{zhang2025generativeverifiersrewardmodeling}, and \emph{pairwise comparison}, where models evaluate pairs of candidates relative to one another~\cite{mahan2024generativerewardmodels}. While pointwise GenRMs are easily compatible with parallel sampling strategies, they are constrained by the inherent limitations of LLMs in verification tasks~\cite{tyen-etal-2024-llms, zhao2025samplescrutinizescaleeffective}. 
Conversely, pairwise GenRMs leverage the comparative strengths of LLMs but extending them to larger candidate pools is non-trivial~\cite{liu2024aligning}. 
% For example, \citet{jiang-etal-2023-llm} perform $\mathcal{O}(N^2)$ pairwise comparisons to rank $N$ samples. 
The computational complexity becomes particularly prohibitive when performing exhaustive pairwise comparisons, requiring $\mathcal{O}(N^2)$ evaluations for ranking $N$ candidates~\cite{jiang-etal-2023-llm}. 
% Even practical 
% Even pa
Approximations such as pairwise knockout tournaments can reduce the computation to $\mathcal{O}(N)$ but increase latency by a factor of $\mathcal{O}(log_{2}N)$~\cite{liu2025pairjudgermperformbestofn}.  
This computational burden is further amplified when using reasoning-based generative reward models for such pairwise comparisons.

% To address the limitations of the current GenRM paradigms, we propose GenSelect, a generative 
% Given that LLMs are 

To address these limitations, we propose GenSelect, where the LLM is tasked with performing the best-of-$N$ judgment given the $N$ candidate solutions. This generalization of GenRM from binary to $N$-ary comparison allows for a far efficient application of reasoning-based LLMs to larger parallel sampling budgets. 
% For example, given $N=16$ solutions, 
To extend GenSelect to even larger sampling budgets, beyond the context window size limitations, we propose a $N$-ary knockout tournament similar to  PairJudge-RM~\cite{liu2025pairjudgermperformbestofn}. 

Our experiments with \texttt{QwQ} and \texttt{DeepSeek-R1} demonstrate that reasoning models are remarkably adept at the GenSelect task out of the box. Our results show that GenSelect substantially outperforms prior approaches on various competition-level math reasoning benchmarks. 
We also find that the GenSelect performance is relatively stable across different inference setups, and thus, $N$-way comparisons with large $N$ allow for highly efficient scaling to higher parallel sampling budgets without performance degradation.  
% Our analysis reveals XYZ...
% Finally, we demonstrate a simple distillation/RL recipe to train a smaller performant GenSelect model. 

% We explore GenSelect in various configurations and find it to be quite s
\section{Related Work}

In this section, we discuss some of the popular reward model work in the context of the mathematical reasoning task. For brevity, we will limit our discussion of reward models for general use cases.
% We divide our discussion into classifier-based re

\subsection{Discriminative Reward Models}
\citet{cobbe2021trainingverifierssolvemath} was one of the first works to use outcome-based reward models (ORMs) where they showed performance benefits with the use of verifiers for selecting the best solution.  
Follow-up work by \citet{yang2024qwen25mathtechnicalreportmathematical} and \citet{liu2025acemathadvancingfrontiermath} demonstrated continued gains with the use of the latest ORMs, even with the dramatic performance improvement of generators.

Besides ORMs, there has been a rich line of work in process reward models (PRMs), where the task is to teach the model to grade based on both the outcome and the process used to arrive at the answer. While earlier work relied on costly human annotations~\cite{lightman2023letsverifystepstep}, recent work utilizes automatic labels via repeated rollouts from different points in the solution trajectory~\cite{zhang2025lessonsdevelopingprocessreward}.

\begin{figure*}
\begin{tcolorbox}[unbreakable,width=\textwidth,colback=white,colframe=NvidiaGreen,title={\centering \large  \textbf{GenRM Prompt}}]
\footnotesize                  
\begin{lstlisting}[
breaklines=true, postbreak={},breakindent=0pt, 
label={lst:genrm-prompt}]
You will be given a challenging math problem followed by a solution. Your task is to systematically analyze this solution to determine whether it is mathematically sound and correct.

Input Format:
Problem: A complex mathematical word problem at advanced high school or college level
Solution: A detailed solution concluding with an answer in \boxed{{}} notation

YOUR TASK

Problem: {problem}

Solution:
{generation}

Evaluation Process:

1. Mathematical Accuracy Check
- Verify all computational steps for arithmetic errors
- Check algebraic manipulations and equation solving
- Validate any formulas, theorems, or mathematical principles used
- Ensure proper mathematical notation and \boxed{{}} format

2. Logical Reasoning Analysis
- Examine the logical progression from problem statement to solution
- Identify any gaps or jumps in reasoning
- Verify that each step follows logically from the previous ones
- Check that the approach appropriately addresses the problem's requirements

3. Completeness and Method Evaluation
- Evaluate handling of edge cases or special conditions if applicable
- Determine if the chosen method is appropriate for the problem type

Your response should include:
1. Step-by-step verification of the mathematical work
2. Analysis of the logical structure and reasoning
3. Identification of any errors, omissions, or weaknesses

End your evaluation with exactly:
Judgment: [Yes/No]
where "Yes" means the solution is mathematically sound and correct, and "No" means the solution contains significant errors or is incorrect.
\end{lstlisting}
\end{tcolorbox}
\caption{Prompt used for GenRM.}
\label{fig:prompt_genrm}
\end{figure*}

\subsection{Generative Reward Models}
Instead of training models to output numerical scores through specialized classification heads, GenRMs leverage the text generation capabilities of large language models by representing correctness decisions using the log probability of specific tokens, typically `Yes' or `No', conditioned on a prompt consisting of the question, corresponding solution, and instruction to judge the correctness of the solution. GenRMs enable two ways of verification/scoring test-time scaling via chain-of-thought reasoning and parallel sampling
~\cite{zhang2025generativeverifiersrewardmodeling}.

Recent work has explored the use of training reasoning models for the task of verification via reinforcement learning (RL). \citet{shi2025heimdalltesttimescalinggenerative} trained such a model for math solution verification, and \citet{guo2025rewardreasoningmodel} trained general verifier models, both models achieving great success with extended reasoning. While the proposed GenSelect method also has a natural RL formulation, we focus on demonstrating the out-of-the-box capabilities of current reasoning models. 

Apart from scoring a standalone solution, GenRMs have also been used and developed for comparing two inputs~\cite{mahan2024generativerewardmodels}. For the math reasoning task, \citet{zhao2025samplescrutinizescaleeffective} note that while \emph {frontier} LLMs can be \emph{weak} verifiers, their capability to identify errors improves when verification is done via comparison of responses. As with \emph{pointwise} GenRMs, long reasoning models trained via RL aid the binary classification task of picking the preferred response as well~\cite{chen2025rmr1rewardmodelingreasoning}. 
While GenRMs, which compare two responses, are an excellent fit for popular reward model benchmarks, using these models for the Best-of-$N$ (BoN) task is non-trivial. Exhaustive pairwise comparisons can be computationally very costly~\cite{jiang-etal-2023-llm}, especially with the new reasoning models, and approximations such as pairwise knockout tournaments can reduce the computation cost at the cost of latency~\cite{liu2025pairjudgermperformbestofn}.

% \subsection{}
\section{Methodologies}

% In this section, we give a brief overview of relevant baselines and contrast them with GenSelect. To make the discussion concrete, we are given an input problem $X$ and a set of corresponding solutions $\{Y_1, \cdots, Y_N\}$. 
% The goal is to select a solution that accurately answers the input problem. 
% We assume access to the function \texttt{Ans($Y$)}, which, given a solution, extracts the final answer from the solution trajectory. 

% Additionally, we assume access to the function \texttt{Summary($Y$)}, which, given a long reasoning solution with backtracking and self-verification, outputs a faithful summary corresponding to the final approach. 
% Initial experiments revealed no benefit from using reasoning traces in comparison to solution summaries in solution representations.
% We therefore use solution summaries for all methods, which also allows compatibility with shorter context window models.
% While reasoning models produce natural summaries (the solution portion that follows the thinking section), our preliminary experiments with solutions generated by \texttt{QwQ} showed slight benefits from generating new summaries using \texttt{Qwen2.5-32B-Instruct} (prompt in Appendix~\ref{sec:soln_summary_prompt}).

In this section, we provide a brief overview of relevant baselines and contrast them with GenSelect. To make the discussion concrete, we consider an input problem $X$ and a set of corresponding solutions $\{Y_1, \cdots, Y_N\}$. 
The goal is to select a solution that accurately answers the input problem. 
We assume access to the function \texttt{Ans($Y$)}, which extracts the final answer from a given solution trajectory. 

Additionally, we utilize the function \texttt{Summary($Y$)}, which produces a faithful summary of a long reasoning solution, including any backtracking and self-verification steps.
Our initial experiments revealed no significant benefit from using complete reasoning traces compared to solution summaries in solution representations.
We therefore adopt solution summaries for all methods, which has the added advantage of compatibility with shorter context window models.
While reasoning models naturally produce summaries (the solution portion following the thinking section), our preliminary experiments with solutions generated by \texttt{QwQ} indicated modest benefits from generating fresh summaries using \texttt{Qwen2.5-32B-Instruct} (prompt details in Appendix~\ref{sec:soln_summary_prompt}).

\subsection{Baselines}

\paragraph{Pass@N.} 
This is the oracle baseline, which selects any of the solutions that reach the ground truth answer. This serves as the upper bound for all the solution ranking methods, including GenSelect.
% the following methodologies, including \textit{GenSelect}.

\paragraph{Majority Voting/Self-Consistency.} Proposed by~\cite{wang2023selfconsistency}, this method selects the most common answer from the solution candidates. \\
$$\text{Majority}(\{\texttt{Ans}(Y_1), \ldots, \texttt{Ans}(Y_N)\})$$

Given that this approach only uses the final answer for aggregation, this is a shallow approach for aggregating multiple responses.

\paragraph{Discriminative Reward Model.} 
A discriminative RM assigns floating-point scores to candidate solutions, which can then be used to select the highest-scoring solution or perform weighted majority voting~\cite{cobbe2021trainingverifierssolvemath, yang2024qwen25mathtechnicalreportmathematical}.
In our experiments, we find that weighted majority voting performs better than selecting the highest-scoring solution, a finding corroborated by \citet{wu2025inferencescalinglawsempirical}.  

As a baseline, we compare against the \texttt{Qwen2.5-Math-RM-72B} model by \citet{yang2024qwen25mathtechnicalreportmathematical}, one of the best-performing outcome-based reward models from prior work.
% Due to \texttt{Qwen2.5-72B-RM}'s limited context window, we evaluate it using solution summaries rather than full reasoning traces. 
% We attempted to establish stronger ORM baselines by fine-tuning Qwen2.5 series models on both complete reasoning solutions and their summaries, but were only able to match \texttt{Qwen2.5-Math-RM-72B}'s performance.

\paragraph{Generative Reward Model.} 
% As a GenRM baseline, we zero-shot prompt the \texttt{QwQ} model with our verification prompt.
% We use this prompt-based GenRM baseline for two reasons. Firstly, GenRM models developed by \citet{zhang2025generativeverifiersrewardmodeling} and \citet{shi2025heimdalltesttimescalinggenerative} have not been publicly released to the best of our knowledge. 
% Secondly, GenSelect can also be used by prompting reasoning models, such as \texttt{QwQ}.
% This allows for a fair comparison between the two paradigms, where GenRM is based on absolute scoring whereas GenSelect is based on comparative scoring. 

We establish a GenRM baseline by zero-shot prompting the \texttt{QwQ} model using our verification prompt (prompt in Figure \ref{fig:prompt_genrm}).
This prompt-based approach serves two purposes: it addresses the unavailability of publicly released GenRM models from prior work \citep{zhang2025generativeverifiersrewardmodeling, shi2025heimdalltesttimescalinggenerative}, while enabling direct comparison with GenSelect under equivalent conditions.
This setup facilitates a fair and meaningful comparison between absolute scoring (GenRM) and comparative scoring (GenSelect) paradigms using the same underlying reasoning model.

% Similar to ORMs, we use GenRMs on solution summaries, as our preliminary experiments suggested that there is no performance benefit to using the reasoning trace.
GenRMs enable straightforward test-time scaling of verification. 
We sample multiple verifications from \texttt{QwQ} at a temperature of 0.6. 

\begin{figure*}
\begin{tcolorbox}[unbreakable,width=\textwidth,colback=white,colframe=NvidiaGreen,title={\centering \large  \textbf{GenSelect Prompt}}]
\footnotesize                  
\begin{lstlisting}[
breaklines=true, postbreak={},breakindent=0pt, 
label={lst:genselect-prompt}]
You will be given a challenging math problem followed by {num_solutions} solutions. Your task is to systematically analyze these solutions to identify the most mathematically sound approach.

Input Format:
Problem: A complex mathematical word problem at advanced high school or college level
Solutions: Detailed solutions indexed 0-{max_idx}, each concluding with an answer in \boxed{{}} notation

YOUR TASK

Problem: {problem}

Solutions:
{solutions}

Evaluation Process:

1. Initial Screening
- Group solutions by their final answers
- Identify and explain mathematical contradictions between different answers
- Eliminate solutions with clear mathematical errors

2. Detailed Analysis
For remaining solutions, evaluate:
- Mathematical precision and accuracy
- Logical progression of steps
- Completeness of mathematical reasoning
- Proper use of mathematical notation, including \boxed{{}}
- Handling of edge cases or special conditions
- For solutions containing and addressing errors, evaluate the error identification and correction methodology.

3. Solution Comparison
Compare viable solutions based on:
- Efficiency of approach
- Clarity of mathematical reasoning
- Sophistication of method
- Robustness of solution (works for all cases)

Your response should include:
1. Brief analysis of conflicting answers
2. Detailed evaluation of mathematically sound solutions
3. Justification for eliminating incorrect solutions
4. Clear explanation for selecting the best approach

End your evaluation with exactly:
Judgment: [IDX]
where IDX is the index 0-{max_idx} of the best solution.
\end{lstlisting}
\end{tcolorbox}
\caption{The prompt used for GenSelect includes 0-indexed solution candidates, and the model must reference the best solution by its corresponding index in the final judgment.}
\label{fig:prompt_genselect}
\end{figure*}

\subsection{GenSelect}
The predominant approach for BoN inference has been to: (a) assign a score to the $N$ solutions, either via a discriminative or generative classifier, and use this score to select the answer, or (b) perform pairwise comparisons and deduce the \emph{best} solution via a sequence of such binary comparisons. Rather than limiting the model to one or two candidate solutions at a time and going a roundabout way to selecting the best candidate, GenSelect tackles the BoN task head-on by directly evaluating all $N$ candidate solutions simultaneously and having the LLM perform an $N$-ary comparison to identify the \emph{best} response (see prompt in Figure~\ref{fig:prompt_genselect}). 
We show that current reasoning models are already adept at this task. 

% In practice, context windows can become an issue for $N > 16$. For example, \texttt{QwQ-32B} has a context window of 40,960 tokens, which includes both prompt and response tokens. To extend GenSelect to even larger sampling budgets, we utilize an $N$-ary knockout tournament similar to PairJudge-RM~\cite{liu2025pairjudgermperformbestofn}. 
% While theoretically the $N$-ary and binary knockout tournaments have the same computational complexity, $O(N)$, and latency, $O(\log N)$, in practice $N$-ary knockout tournaments are much more efficient. 
% For example, given 64 candidate solutions, binary comparisons require six rounds and a total of 63 comparisons, while 16-way comparisons require only two rounds and a total of 5 comparisons.
% This $N$-ary knockout is much more

In practice, context window limitations present scalability challenges for $N > 16$. For instance, \texttt{QwQ-32B} has a maximum context of 40,960 tokens, which includes both the prompt and response tokens. To scale GenSelect to larger sampling budgets, we employ an $N$-ary knockout tournament approach, following the methodology established in PairJudge-RM~\cite{liu2025pairjudgermperformbestofn}.
Although $N$-ary and binary knockout tournaments maintain equivalent theoretical computational complexity of $\Theta(N)$ and latency of $\Theta(log N)$, $N$-ary tournaments demonstrate substantially superior practical efficiency. For example, when evaluating 64 candidate solutions, binary comparisons require six sequential rounds comprising 63 total comparisons, whereas 16-way comparisons require merely two rounds with five total comparisons, representing a significant reduction in computational overhead. This becomes even more evident with the overhead of reasoning models performing long reasoning to determine the best solution in each comparison. 
We also demonstrate that GenSelect exhibits stable performance across various values of $N$, indicating that efficiency gains can be achieved with larger $N$ without performance degradation.

\begin{table}[t]
\centering
\renewcommand{\arraystretch}{1.2} % Increase vertical padding
\begin{tabular}{lc}
\toprule
\textbf{Problem source} & \textbf{\# of Problems} \\
\midrule
AIME 2024 & \phantom{1}30 \\
AIME 2025 & \phantom{1}30 \\
HMMT Nov 2024 & \phantom{1}62 \\
HMMT Feb 2024 & \phantom{1}68 \\
HMMT Feb 2025 & \phantom{1}66 \\
\midrule
\textbf{Total} & 256 \\
\bottomrule
\end{tabular}
\caption{Composition of Comp-Math-24-25.}
\label{tab:validation_dataset}
\end{table}

\begin{table*}[t]
\centering
\renewcommand{\arraystretch}{1.2} % Increase vertical padding
\setlength{\tabcolsep}{4pt}
\begin{tabular}{lcccc}
\toprule
& \multicolumn{2}{c}{\textbf{DeepSeek-R1-0528}} & \multicolumn{2}{c}{\textbf{QwQ-32B}} \\
\cmidrule(lr){2-3} \cmidrule(lr){4-5}
\textbf{$N$} & \textbf{Generation ($2N$)} & \textbf{Generation ($N$) + GenSelect ($N$)} & \textbf{Generation ($2N$)} & \textbf{Generation ($N$) + GenSelect ($N$)} \\
\midrule
4 & \textbf{82.8} & 82.4 & 66.4 & \textbf{69.5} \\ 
8 & 84.4 & \textbf{85.9} & 66.8 & \textbf{69.5} \\
16 & 84.4 & \textbf{87.1} & 68.0 & \textbf{71.9} \\
\bottomrule
\end{tabular}
\caption{Comparison of performance with \texttt{DeepSeek-R1-0528} and \texttt{QwQ-32B} when spending inference compute budget entirely on solution generation, followed by majority voting vs dividing the inference budget equally into solution and GenSelect generation. }
\label{tab:inference-budget-combined}
\end{table*}

\section{Experimental Setup and Results}

\subsection{Evaluation Benchmark}
The AIME competitions are a popular benchmark for math reasoning, but they consist of only 30 questions, resulting in high variance in performance measurements. To reduce this variance, we combine problems from the American Invitational Mathematics Examinations (AIME) and Harvard-MIT Mathematics Tournaments (HMMT) for the years 2024 and 2025. 
We exclude proof-based questions and those awarding partial credit based on estimated accuracy. 
We refer to this dataset as \texttt{Comp-Math-24-25}, which consists of 256 problems, as detailed in Table \ref{tab:validation_dataset}.

\begin{table}[h!]
\centering
\renewcommand{\arraystretch}{1.2} % Increase vertical padding
\begin{tabular}{lc}
\toprule
\textbf{BoN Method} & \textbf{Accuracy (in \%)} \\
\midrule
Pass@64    &  85.2  \\
\midrule
Majority@64 & 68.4 \\
Qwen2.5-Math-RM-72B & 66.8 \\
\midrule
QwQ GenRM & 69.1\\ 
QwQ GenSelect@1 & \textbf{72.1} \\
QwQ GenSelect@8 & \textbf{73.4} \\
\bottomrule
\end{tabular}
\caption{Accuracy on Comp-Math-24-25 for solutions generated by \texttt{QwQ-32B}. 
For both GenRM and GenSelect, we use the \texttt{QwQ-32B} model itself to score the candidate solutions. 
For GenSelect, we conduct a 16-way competition followed by a 4-way competition to determine the best solution.
}
\label{tab:training-comparison}
\end{table}

\subsection{Evaluation Details}
We evaluate two of the most popular open-weight reasoning models, namely, \texttt{QwQ-32B} and \texttt{DeepSeek-R1-0528}, in our experiments. For both models, we use a sampling temperature of 0.6 to generate solutions, and when the model is used as a verifier/reward model. 

\subsection{Results}

\paragraph{Comparison with Baselines.} 

Table~\ref{tab:training-comparison} presents a comprehensive comparison of various Best-of-N selection methods alongside our proposed GenSelect approach. The underperformance of \texttt{Qwen2.5-Math-RM-72B} relative to majority voting can be attributed to distribution mismatch, as the model was trained on a different data distribution. Additionally, prior research has demonstrated that discriminative reward models tend to degrade relative to majority voting at large sampling budgets. While QwQ GenRM achieves modest improvements over the majority baseline, these gains are substantially exceeded by QwQ GenSelect@1, which demonstrates that leveraging the comparative strengths of LLMs for selecting the best solution is more effective than scoring each solution individually and then selecting the best one. The performance advantage becomes even more pronounced when the GenSelect pipeline is repeated eight times with majority voting applied to the resulting outputs, achieving 73.4\% accuracy. 
This substantial improvement underscores that current reasoning models, like \texttt{QwQ}, are adept at GenSelect with simple prompting.

% efficacy of GenSelect, and its applicability with 
% Table~\ref{tab:training-comparison} compares the different BoN methods, along with the proposed GenSelect method. 
% The \texttt{Qwen2.5-Math-RM-72B} doing worse than majority could be because the model was trained on different data distribution and it has also been shown in prior work that discriminative reward models start underperforming majority at large sampling budgets.  
% The QwQ GenRM improves marginally over the majority baseline, but pales in comparison to the gains of 
% QwQ GenSelect@1.  
% This suggests that leveraging the comparative strengths of LLMs for selecting the best solution is better than scoring each solution individually and then selecting the maximum.  
% Furthermore, GenSelect improves further to 73.4 when the GenSelect pipeline is repeated 8 times and majority is performed on the these 8 outputs. 
% significantly outperforms the majority voting baseline (68.4\%) with further gains with QwQ GenSelect@8. 

% Notably, GenSelect achieves 73.4\% accuracy with only 8 evaluations, surpassing QwQ GenRM@32's performance of 69.1\% despite using four times fewer . 
% This efficiency gain is particularly striking given that GenSelect employs a tournament-based approach (16-way followed by 4-way competition) that requires significantly less computational overhead while maintaining superior selection quality. The results highlight GenSelect's ability to achieve strong performance with reduced computational cost compared to both traditional reward models and repeated generative scoring approaches.

\begin{table}[h!]
\centering
\renewcommand{\arraystretch}{1.2} % Increase vertical padding
\begin{tabular}{lcc}
\toprule
\textbf{$N$} & \textbf{GenSelect@1}  & \textbf{GenSelect@8} \\
\midrule
2 &  72.1 & 73.4 \\ 
4 & 72.6 & 73.0 \\
8 &  72.3 & 73.4 \\
16 & 72.1 & 73.4\\
\bottomrule
\end{tabular}
\caption{Comparison of GenSelect performance with different values of $N$ which determines the $N$-ary comparisons performed. 
}
\label{tab:inference-stability}
\end{table}

\paragraph{Stability of GenSelect.} 
In this ablation, we compare the performance of GenSelect in different inference setups. In particular, we change the value of $N$ from $\{2, 4, 8, 16\}$. Note that when $N=2$, for a candidate solution set of 64 solutions, we require six rounds of comparison to decide the best solution, whereas just two rounds of comparisons are required with $N=8$ and $16$.

Table~\ref{tab:inference-stability} presents the performance of these different setups, and from the results, it is pretty evident that GenSelect is relatively stable across the different inference setups. This suggests that we can make $N$ large (up to context window limits), and accelerate the best solution selection pipeline without performance degradation.

\paragraph{Generation vs Verification.}
Given an inference compute budget, are we better off spending it on generating more solutions, or spending some of it on verification, or in our case, on GenSelect. 
To answer this question, we compare two scenarios: (a) $2N$ solution generations followed by majority voting, and (b) $N$ solution generations followed by $N$ GenSelect generations. Note that while the two systems may use a similar amount of computation, the latency of the second one would most likely be worse than the first one, since GenSelect pipeline can only start once the solution generations have finished.

% \begin{table}[t!]
% \centering
% \renewcommand{\arraystretch}{1.2} % Increase vertical padding
% \setlength{\tabcolsep}{2pt}
% \begin{tabular}{lcccc}
% \toprule

% \textbf{$N$} & \textbf{Generation ($2N$)}  & \textbf{Generation ($N$) + GenSelect ($N$)} \\
% \midrule
% 4 & \textbf{82.8} & 82.4 \\ 
% 8 & 84.4 & \textbf{85.9} \\
% 16 & 84.4 & \textbf{87.1} \\
% \bottomrule
% \end{tabular}
% \caption{Comparison of performance with \texttt{DeepSeek-R1-0528} when spending inference compute budget entirely on solution generation vs dividing the inference budget equally into solution and GenSelect generation. 
% }
% \label{tab:inference-budget-deepseek}
% \end{table}

% \begin{table}[t]
% \centering
% \renewcommand{\arraystretch}{1.2} % Increase vertical padding
% \setlength{\tabcolsep}{2pt}
% \begin{tabular}{lcccc}
% \toprule

% \textbf{$N$} & \textbf{Generation ($2N$)}  & \textbf{Generation ($N$) + GenSelect ($N$)} \\
% \midrule
% 4 & 66.4 & \textbf{69.5} \\ 
% 8 & 66.8 &  \textbf{69.5} \\
% 16 & 68.0 & \textbf{71.9} \\
% \bottomrule
% \end{tabular}
% \caption{Comparison of performance with \texttt{QwQ-32B} when spending inference compute budget entirely on solution generation vs dividing the inference budget equally into solution and GenSelect generation. 
% }
% \label{tab:inference-budget-qwq}
% \end{table}

Table~\ref{tab:inference-budget-combined} presents results for this inference compute allocation comparison for \texttt{DeepSeek-R1-0528} and \texttt{QwQ-32B}. For the stronger model, \texttt{DeepSeek-R1-0528}, we see that focusing on generation-only for lower sampling budgets is preferred, but for $N >= 8$, allocating equal inference compute to GenSelect is preferred. For the weaker \texttt{QwQ-32B} model, allocating equal compute to GenSelect is advantageous for all sampling budgets in our study.

\begin{table*}[t]
\centering
\renewcommand{\arraystretch}{1.2} % Increase vertical padding
\begin{tabular}{lcccc}
\toprule
\multirow{2}{*}{\textbf{Model}} & \multicolumn{4}{c}{\textbf{Comp-Math-24-25}} \\
\cmidrule(lr){2-5}
 & \textbf{AIME24} & \textbf{AIME25} & \textbf{HMMT-24-25}  & \textbf{All}\\
\midrule
DeepSeek-R1-0528 & 85.9 (93.3) & 80.5 (86.7) & 67.5 (82.1)  & 71.2 (84.0) \\
\quad + Self GenSelect@32 & 93.3 & 90.0 & 85.7 & 87.1 \\
\midrule
QwQ-32B  &  78.1 (86.7) &  67.2 (76.7) &  56.1 (64.3)  &  60.0 (68.4) \\
\quad + Self GenSelect@32 & 90.0 & 73.3 & 70.4 & 73.0 \\
\bottomrule
\end{tabular}
\caption{All models are evaluated with a maximum of 32K output tokens, temperature of $0.6$, and top-p $0.95$. We present metrics as pass@1 (maj@64) where pass@1 is an average accuracy across 64 generations and maj@64 is the result of majority voting. For HMMT, we use the LLM-judge setup of~\cite{toshniwal2024openmathinstruct} to verify the answers. For GenSelect, we use $N=8$, which requires two rounds of scoring eight solutions each. We repeat GenSelect 32 times with random solution permutations and perform majority voting over the answers selected by GenSelect. }
\label{tab:final-results}
\end{table*}

\paragraph{Final Results.}
Table~\ref{tab:final-results} presents the results for both \texttt{DeepSeek-R1-0528} and \texttt{QwQ-32B}, along with their Self-GenSelect counterparts. For both models, we see a significant improvement in performance over the majority baseline. In particular, we see the gains are predominantly in the HMMT-24-25 split of Comp-Math-24-25.

\section{Conclusion}

In this work, we introduced GenSelect, a tournament-based Best-of-N selection method that leverages the comparative evaluation capabilities of large language models. Our experimental evaluations for mathematical reasoning tasks demonstrate that GenSelect consistently outperforms established baselines including majority voting, discriminative reward models, and generative reward models. The method exhibits remarkable stability across different tournament configurations and addresses practical scalability limitations through efficient $N$-ary knockout tournaments that respect context window constraints while significantly reducing computational overhead.

Our analysis of inference compute allocation reveals that GenSelect's effectiveness varies with model capability, becoming advantageous for stronger models at moderate sampling budgets while benefiting weaker models across all tested configurations. The simplicity of GenSelect's implementation---requiring only straightforward prompting without specialized training---combined with its consistent performance gains, makes it an immediately applicable technique for enhancing mathematical reasoning systems. Future work could explore extending GenSelect to other reasoning domains and general tasks. The GenSelect formulation is also suitable for reinforcement learning, and future work can leverage RL to learn the comparative capabilities of larger reasoning language models in smaller LLMs.

% \section*{Impact Statement}


\bibliography{main}
\bibliographystyle{icml2025}

%%%%%%%%%%%%%%%%%%%%%%%%%%%%%%%%%%%%%%%%%%%%%%%%%%%%%%%%%%%%%%%%%%%%%%%%%%%%%%%
%%%%%%%%%%%%%%%%%%%%%%%%%%%%%%%%%%%%%%%%%%%%%%%%%%%%%%%%%%%%%%%%%%%%%%%%%%%%%%%
% APPENDIX
%%%%%%%%%%%%%%%%%%%%%%%%%%%%%%%%%%%%%%%%%%%%%%%%%%%%%%%%%%%%%%%%%%%%%%%%%%%%%%%
%%%%%%%%%%%%%%%%%%%%%%%%%%%%%%%%%%%%%%%%%%%%%%%%%%%%%%%%%%%%%%%%%%%%%%%%%%%%%%%
\newpage
\appendix
\onecolumn

\section{Prompts}

\subsection{Solution Summary Prompt}
\label{sec:soln_summary_prompt}
\begin{tcolorbox}[breakable,width=\textwidth,colback=white,colframe=NvidiaGreen,title={\centering \large  \textbf{Summary Prompt}}]
\footnotesize                  
\begin{lstlisting}[
breaklines=true, postbreak={},breakindent=0pt, 
label={lst:summary-prompt}]
I will give you a math problem and a long solution to that problem exploring different approaches, making mistakes along the way, correcting them, switching around and so on. But eventually that solution gets to the right approach and solves the problem. Your task is to write a clean version of the final correct solution without all the exploration. Cover all the details of the final solution.

Problem:
{problem}

Solution:
{generation}

Now write a clean version of the final correct solution without all the exploration but cover all the details of the final solution. 

\end{lstlisting}
\end{tcolorbox}

% \subsection{GenRM Prompt}
% \label{sec:genrm_prompt}
% \begin{tcolorbox}[breakable,width=\textwidth,colback=white,colframe=NvidiaGreen,title={\centering \large  \textbf{GenRM Prompt}}]
% \footnotesize                  
% \lstinputlisting[
% breaklines=true, postbreak={},breakindent=0pt, 
% label={lst:genrm-prompt}]{prompts/genrm.md}
% \end{tcolorbox}

\end{document}